\documentclass[11pt,letterpaper,twocolumn]{article}
\usepackage{times}
\usepackage{latexsym}
\usepackage{framed}
\usepackage{bm}

\usepackage{graphicx} 

\usepackage{natbib}

\usepackage{algorithm}
\usepackage{algorithmic}
\usepackage[algo2e]{algorithm2e}

\usepackage{amsmath}
\usepackage{amsfonts}

\title{Self-organized Hierarchical Softmax}

\author{Yikang Shen \\ University of Montreal \\ {\tt yi-kang.shen@umontreal.ca} \and Shawn Tan \\University of Montreal \\ {\tt tanjings@iro.umontreal.ca} \and Christopher Pal \\ Polytechnique Montreal \\ {\tt christopher.pal@polymtl.ca} \and Aaron Courville \\ University of Montreal \\ {\tt aaron.courville@gmail.com}}

\date{}

\begin{document} 

\maketitle

\begin{abstract} 
We propose a new self-organizing hierarchical softmax formulation for neural-network-based language models over large vocabularies. Instead of using a predefined hierarchical structure, our approach is capable of learning word clusters with clear syntactical and semantic meaning during the language model training process. We provide experiments on standard benchmarks for language modeling and sentence compression tasks. We find that this approach is as fast as other efficient softmax approximations, while achieving comparable or even better performance relative to similar full softmax models.
\end{abstract} 

\section{Introduction}
The softmax function and its variants are an essential part of neural network based models for natural language tasks, such as language modeling, sentence summarization, machine translation and language generation.

Given a hidden vector, the softmax can assign probability mass to each word in a vocabulary. The hidden vector could be generated from the preceding context, source sentence, dialogue context, or just random variables. The model decides how the context is converted into the hidden vector, and there are several choices for this, including recurrent neural network \cite{hochreiter1997long, mikolov2010recurrent}, feed forward neural network \cite{bengio2003neural} or log-bilinear models \cite{mnih2009scalable}. In our experiments here, we use a long short-term memory (LSTM) model for language modeling, and a sequence-to-sequence model with an attention mechanism for sentence compression. Both models are simple but have been shown capable of achieving state-of-the-art results. Our focus is to demonstrate that with a well designed structure, the hierarchical softmax approach can perform as accurately as the full softmax, while maintaining improvements in efficiency.

For word-level models, the size of the vocabulary is very important for higher recall and a more accurate understanding of the input. However the training speed for models with softmax output layers quickly decreases as the vocabulary size grows. This is due to the linear increase of parameter size and computation cost with respect to vocabulary. 

Many approaches have been proposed to reduce the computational complexity of large softmax layers \cite{mikolov2011strategies, chen2015strategies, grave2016efficient}. These methods can largely be divided into two categories:
\begin{enumerate}
\item Approaches that can compute a normalized distribution over the entire vocabulary with a lower computational cost \cite{chen2015strategies, grave2016efficient}. Normalized probabilities can be useful for sentence generation tasks, such as machine translation and summarization. 

\item Methods that provide unnormalized values \cite{bengio2003quick, mikolov2013distributed}. These methods are usually more efficient in the training process, but less accurate. 
\end{enumerate}

In this paper, we propose a self-organized hierarchical softmax, which belongs in the first category. In contrast to previous hierarchical softmax methods which have used predefined clusters, we conjecture here that a hierarchical structure learned from the corpus may improve model performance. Instead of using term frequencies as clustering criteria \cite{mikolov2011strategies, grave2016efficient}, we want to explore the probability of clustering words together considering their preceding context. The main contributions of this paper are as follows:
\begin{itemize}
\item We propose an algorithm to learn a hierarchical structure during the language model learning process. The goal of this algorithm is to maximize the probability of a word belonging to its cluster considering its preceding context.
\item We conduct experiments for two different tasks: language modeling and sentence summarization. Results show
that our learned hierarchical softmax can achieve comparable accuracy for language modeling, and even better performance for summarization when compared to a standard softmax. We also provide clustering results, which indicate a clear semantic relevance between words in the same cluster.
\item Empirical results show that our approach provides a more than a $3\times$ speed up compared to the standard softmax.
\end{itemize} 

\section{Related Work}
Representing probability distributions over large vocabularies is computationally challenging. In neural language modeling, the standard approach is to use a softmax function that output a probability vector over the entire vocabulary. Many methods have been proposed to approximate the softmax with lower computational cost \cite{mikolov2011strategies,chen2015strategies,bengio2003quick,grave2016efficient}. We briefly review the most popular methods below. 

\subsection{Softmax-based approaches}
\paragraph{Hierarchical Softmax (HSM):} \cite{goodman2001classes} and its variants are the most popular approximations. In general, this approach organizes the output vocabulary into a tree where the leaves are words and intermediate nodes are latent variables, or classes. The tree structure could have many levels and there is a unique path from root to each word. The probability of a word is the product of probabilities of each node along its path. In practice, we could use a tree with two layers, where we want to organize words into simple clusters. In this case, the computational complexity reduces from $O(|V|)$ to $O(\sqrt{|V|})$. If we use a deeper structure like the Huffman Tree, the computational complexity could be reduced to $O(\log|V|)$. In general, the hierarchical structure is built on frequency binning \cite{mikolov2011extensions,grave2016efficient} or word similarities \cite{chelba2013one, le2011structured,chen2015strategies}. In this paper, we propose another word-similarity-based hierarchical structure. But, instead of performing k-means over pre-learned word embeddings, we propose a new approach that learns hierarchical structure based on the model's historical prediction during the language model learning process. 

\paragraph{Differentiated softmax (D-softmax):} \cite{chen2015strategies} is based on the intuition that not all words require the same number of parameters: The many occurrences of frequent words allows us to fit many parameters to them, while extremely rare words might only allow us to fit relatively few parameters. D-softmax assign different dimension of vector to words according to their frequency to speed up the training and save memory.
Adaptive softmax \cite{grave2016efficient} can be seen as a combination of frequency binning HSM and 
D-softmax. 

\paragraph{CNN-softmax:} \cite{jozefowicz2016exploring} is inspired by the idea that we could use convolution network to produce word embedding from a character level model. Aside from a big reduction in number of parameters and incorporating morphological knowledge from words, this method can also easily deal with out-of-vocabulary words, and allows parallel training over corpora that have different vocabulary size.
But this method does not decrease the computational complexity compared to the standard full softmax \cite{jozefowicz2016exploring}.

\subsection{Sampling-based approaches}
Sampling based approaches approximate the normalization in the denominator of the softmax with some other loss that is cheap to compute. However, sampling based approaches are only useful at training time. During inference, the full softmax still needs to be computed to obtain a normalized probability. These approaches have been successfully applied to language modeling \cite{bengio2008adaptive}, machine translation \cite{Jean2015On}, and computer vision \cite{joulin2016learning}.

\paragraph{Importance sampling (IS):} \cite{bengio2003quick,bengio2008adaptive} select a subset of the vocabulary as negative samples to approximate the softmax normalization. 
Originally unigram or bigram distribution of word in entire corpus are used for sampling negative samples \cite{bengio2003quick}, but researchers found that sampling from a more carefully designed distribution could help achieve a better accuracy.
Instead, two variants of n-gram distributions are proposed:
\begin{enumerate}
\item an interpolated bigram distribution and unigram distribution \cite{bengio2008adaptive}, 
\item a power-raised unigram distribution \cite{mikolov2013efficient}.
\end{enumerate}

\paragraph{Noise Contrastive Estimation (NCE):} Noise Contrastive Estimation (NCE) is proposed in \cite{gutmann2010noise, mnih2013learning} as a more stable sampling method than IS. NCE does not try to estimate the probability of a word directly. Instead, it uses an auxiliary loss that works to distinguish the original distribution from a noisy one. \cite{mnih2012fast} showed that good performance can be achieved even without computing the softmax normalization.

\section{Self-organized Hierarchical Softmax}
\subsection{Cluster-based Hierarchical Softmax}
We employ a modified 2-layer hierarchical softmax to compute the distribution of next word in a sentence. Given vocabulary $\mathbf{V}$ of size $N$, and pre-softmax hidden states $\mathbf{h}$, we first project $\mathbf{h}$ into a cluster vector $\mathbf{h}_c$ and a word vector $\mathbf{h}_w$,
\begin{equation}
\begin{bmatrix}
\mathbf{h}_c \\ 
\mathbf{h}_w
\end{bmatrix}
=\text{Relu}
\begin{bmatrix}
\mathbf{W}_c \mathbf{h} \\ 
\mathbf{W}_w \mathbf{h}
\end{bmatrix}
\end{equation}
where $\mathbf{W}_c, \mathbf{W}_w \in \mathbb{R}^{d \times d}$. The cluster distribution can be expressed as
\begin{equation}
	P(c|\mathbf {h} )={\frac {\exp \left( \mathbf {h}_c ^{\mathsf {T}}\mathbf {U}^\mathcal{C}_{c} \right) }{\sum _{c' \in \mathcal{C}} \exp \left( \mathbf {h}_c ^{\mathsf {T}}\mathbf {U}^\mathcal{C}_{c'} \right) }}
    \label{classprob}
\end{equation}
where $\mathcal{C}$ is set of clusters, $\mathbf{U}^\mathcal{C} \in \mathbb{R}^{|\mathcal{C}| \times d}$ is vector representation of clusters.
The in-cluster probability function is
\begin{equation}
	P(w_t|\mathbf {h}, \mathcal{C}(w_t))={\frac {\exp \left( \mathbf {h}_w ^{\mathsf {T} }\mathbf {U}^\mathbf{V}_{w_t} \right) }{\sum _{w' \in \mathcal{C}(w_i)} \exp \left( \mathbf {h}_w ^{\mathsf {T}}\mathbf {U}^\mathbf{V} _{w'} \right) }}
    \label{inclassprob}
\end{equation}
where $\mathbf {U}^\mathbf{V} \in \mathbb{R}^{|\mathbf{V}| \times d}$ is vector representation of words, 
$\mathcal{C}(w_t)$ is the cluster assigned to $w_t$ in $\mathcal{C}$.
Thus, the final probability function is
\begin{equation}
	P(w_t|\mathbf {h}) = P(\mathcal{C}(w_t)|\mathbf {h} ) P(w_t|\mathbf {h}, \mathcal{C}(w_t) )
\end{equation}
If the number of cluster is in $O(\sqrt{N})$ and the maximum number of words in cluster is in $O(\sqrt{N})$, then the computational cost of normalization at each layers is only $O(\sqrt{N})$, (as opposed to $O(N)$ for the standard softmax).  Thus a large matrix dot product is transformed into two small matrix dot product, which are very efficient on a GPU \cite{grave2016efficient}.

\subsection{Cluster Perplexity}
In order to evaluate the quality of a clustering over words, we propose the cluster perplexity:
\begin{equation}
\text{ppl}_{\mathrm{cluster}}(\mathcal{C}) = 2^{\frac{1}{M} \sum_{w_t} - \log_2 p(\mathcal{C}(w_t) | \mathbf{w}_{<t})}
\label{eq:cluster}
\end{equation}
where $M$ is number of words in the dataset, $\mathbf{w}_{<t}$ is context preceding $w_t$, $p(\mathcal{C}(w_t) | \mathbf{w}_{<t})$ is the probability that words in cluster $\mathcal{C}(w_t)$ appear behind $\mathbf{w}_{<t}$. Given a word cluster $\mathcal{C}$ and $\mathbf{w}_{<t}$, this metric evaluate the difficulty to choose correct cluster. If words that share similar context have been successfully grouped together, the $\text{ppl}_{\mathrm{cluster}}$ should be small.

In addition to $\text{ppl}_{cluster}$, we also propose the in-cluster perplexity:
\begin{equation}
\text{ppl}_{\mathrm{in-cluster}} ( \mathcal{C} ) = 2^{\frac{1}{M} \sum - \log_2 p(w_t | \mathbf{w}_{<t}, \mathcal{C}(w_t))}
\label{eq:incluster}
\end{equation}
where $p(w_t | \mathbf{w}_{<t}, \mathcal{C}(w_t))$ is the probability of word $w_t$ appearing after $\mathbf{w}_{<t}$ given a subset of vocabulary $\mathcal{C}(w_t)$. If $\mathcal{C}(w_t)$ contains words that share the same context with $w_t$, $\text{ppl}_{\mathrm{in-cluster}}$ should be large.

\subsection{Optimizing Cluster Perplexity}
With the definitions in Equations \ref{eq:cluster} and 
\ref{eq:incluster} established, our goal is to minimize $\text{ppl}_{\mathrm{cluster}}(\mathcal{C})$:
\begin{align}
&\text{argmin}_{\mathcal{C}} \text{ppl}_{\mathrm{cluster}}(\mathcal{C}) \nonumber \\ 
&= - \text{argmax}_{\mathcal{C}} \frac{1}{M} \sum_{w_t} \log_2 p(\mathcal{C}(w_t) | \mathbf{w}_{<t}) \nonumber\\
&= - \text{argmax}_{\mathcal{C}} \sum_{w \in \mathbf{V}} \frac{n_w}{M} \frac{\sum_{w_t = w} \log_2 p(\mathcal{C}(w) | \mathbf{w}_{<t})}{n_w}  \nonumber\\
&= - \text{argmax}_{\mathcal{C}} \sum_{w \in \mathbf{V}} tf(w) g(w,\mathcal{C}(w)) \label{weightedsum}
\end{align}
where $\mathrm{tf}(w)=\frac{n_w}{M}$ is term frequency of word $w$ in the corpus, and
\begin{equation}
g(w,\mathcal{C}(w))=\frac{1}{n_w} \sum_{w_t = w} \log_2 p(\mathcal{C}(w) | \mathbf{w}_{<t})
\end{equation}
is the average of $\log_2 p(\mathcal{C}(w) | \mathbf{w}_{<t})$ over different preceding contexts $\mathbf{w}_{<t}$ that followed by word $w$.

According to equation \ref{weightedsum}, we need a $\mathcal{C}$ that maximize the weighted sum of $g(w,\mathcal{C}(w))$. While directly computing $g(w,\mathcal{C}(w))$ is intractable, 
the output of equation \ref{classprob} at training time can be considered a sample of $p(c | \mathbf{w}_{<t})$. We propose to use exponential smoothing to estimate $g(w,\mathcal{C}(w))$:
\begin{align}
	q_{\tau}(w_t,c) = & \lambda(w_t) q_{\tau-1}(w_t,c) \nonumber \\
    &+ (1 - \lambda(w_t)) \log_2 P(c|\mathbf{h}_t)
    \label{classprob_update}
\end{align}
$q_{\tau}(c|w_t)$ is a weighted sum over all historical samples of $p(c|\mathbf{w}_{<t})$ under different context and parameters, previously seen in training. 
The smoothing factor $\lambda(w_t)$ is defined as
\begin{equation}
\lambda(w_t) = \frac{1}{f_{w_t}}
\end{equation}
where $f_{w_t}$ is the raw count of $w_t$ in the entire dataset.

\subsection{Greedy Word Clustering}
In practice, we assigned two constraints to each cluster:
\begin{enumerate}
\item The number of words in each cluster cannot exceed $\gamma \sqrt{N}$, where 
$\gamma > 1$ is a hyperparameter; and 
\item The sum of term frequencies in each cluster should be smaller than a frequency budget $f_b$. This is known as the the frequency-budget trick \cite{chen2015strategies}. 
\end{enumerate}
These constraints prevent us from getting clusters that are either too large, which make computing in-cluster distributions very expensive, or too unbalanced in frequency, which will bias our word cluster distribution.

\begin{algorithm}[t]
	\KwData{Word cluster distribution $q(c|w)$}
	\KwResult{New clusters $\mathcal{C}_t$}
    
    Generate $\sqrt{N}$ empty clusters
    
	\For{word $w$ in $tf(w)$ descend order}{
    	\For{cluster $c$ in $q(w,c)$ descend order}{
			\If {$|c| < \gamma \sqrt{N}$ and $\sum_{w \in c} tf_w < f_{b}$} {
            	add $w$ into $c$
                
                break
            }
        }
	}
	\Return clusters
	\caption{Greedy Word Cluster Assignment}
    \label{Estep}
\end{algorithm}
The greedy algorithm \ref{Estep} is proposed to optimize the cluster perplexity. As each cluster has limited positions for words, some words $w$ cannot be assign to their best cluster $c = \text{argmax}_c q(w,c)$. If we assign words according to certain order, then words at the tail end of the sequence will be less likely to be assign to their best cluster. In the algorithm, we assign words to clusters in descending order of their term frequency $tf(w)$. In this schema, high frequency words $w$ have priority to choose clusters, because they have higher weight in equation \ref{weightedsum}.

\subsection{Training Language Model with Self-organized HSM}
In the training phase, we start from a randomly initialized word cluster, and update parameters using gradient descent based optimization algorithms, updating word cluster every $K$ iterations. $K$ is a hyperparameter that is chosen based on dataset size and vocabulary size. 
This learning process can also be considered as an EM algorithm: In the E-Step, we update the clusters; in M-Step, we update parameters based on the new clusters.

\section{Language Modeling Experiment}
Language Modeling (LM) is a central task in NLP. The goal of LM is to learn a probability distribution over a sequence of tokens from a given vocabulary set $\mathbf{V}$. The joint distribution is defined as a product of conditional distribution of tokens given their preceding context. Given a sequence of word $w_1,...,w_T \in \mathbf{V}$, the probability distribution can be defined as:
\begin{equation}
P(w_1,...,w_T) = \prod_{t=1}^{T} P(w_t|w_1,...,w_{t-1})
\end{equation}
To address this problem, much work has been done on both parametric and non-parametric approaches. In recent years, parametric models based on neural networks have became the standard method. In our experiment, we used the standard word-level Long-Short Term Memory (LSTM) model, since multiple works show it can obtain state-of-the-art performance on different datasets \cite{jozefowicz2016exploring, grave2016efficient}.

\subsection{Dataset}
We evaluate our method on the text8\footnote{http://mattmahoney.net/dc/textdata} dataset, and use the perplexity (ppl) as an evaluation metric. We also provide the training time for full softmax and our approach. Text8 is a standard compression dataset containing a pre-processed version of the first 100 million characters from Wikipedia in English. It has been recently used for language modeling (Mikolov et al., 2014) and has a vocabulary of 44k words. The dataset partitioned into a training set (first 99M characters) and a development set (last 1M characters) that is used to report performance \cite{mikolov2014learning}.

\subsection{Implementation}
In our experiments, we use the same setting as the one reported in  \cite{grave2016efficient}. A one-layer LSTM model is used. Both the dimension of hidden state and dimension of the input word embeddings is set to 512. LSTM parameters are regularized with weight decay ($\lambda=10^{-6}$). Batch size is set to 128. We use Adagrad \cite{duchi2011adaptive} with learning rate 0.1, the norm of the gradients is clipped to 0.25, and a 20 steps gradient truncation is applied. 

For our model, we set the number of clusters to $\sqrt{\mathbf{V}}$, and the maximum number of words in each cluster is $\gamma \sqrt{\mathbf{V}}$ with $\gamma=1.5$, and frequency budget $f_b$ is $0.1$ as in \cite{chen2015strategies}. We update the word clusters every $K=1000$ mini-batches.

\subsection{Baseline Methods}
We compare the proposed approach with (1) full softmax, (2) importance sampling \cite{bengio2003quick}, (3) hierarchical softmax (HSM) with frequency binning \cite{mikolov2011strategies}, (4) differentiated softmax \cite{chen2015strategies}, and (5) adaptive softmax \cite{grave2016efficient}. As we use the same implementation settings in \cite{grave2016efficient}, we use their experiment results for baseline methods. Instead of using torch, we use theano \cite{bergstra2010theano} to implement our approach. Thus, in order to compare computation time, we implement another full softmax language model with theano. Our full softmax has the same perplexity on the development set as the one reported in \cite{grave2016efficient}.

\subsection{Experimental results}
\begin{table}[t]
	\center
	\caption{Experiment results on Text8 dataset, Adagrad after 5 epochs, learning rate is 0.1}
    \label{text8_result}
	\begin{tabular}{c|c}
		& ppl \\
		\hline
		Full softmax & 144 \\
        Importance sampling & 166 \\
        HSM (frequency binning) & 166 \\
        D-softmax & 195 \\
        Adaptive softmax & 147 \\
        \hline
        Our full softmax & 144.17 \\
        Self-organized HSM & 144.77
	\end{tabular}
\end{table}

\begin{table}[t]
	\center
	\caption{Training time on Text8 training set}
    \label{text8_time}
	\begin{tabular}{c|c}
	& Training time \\
    \hline
    Our full softmax & 77 min \\
    Self-organized HSM & 20 min
	\end{tabular}
\end{table}

\begin{figure}[!ht]
	\centering
	\includegraphics[width=0.95\columnwidth]{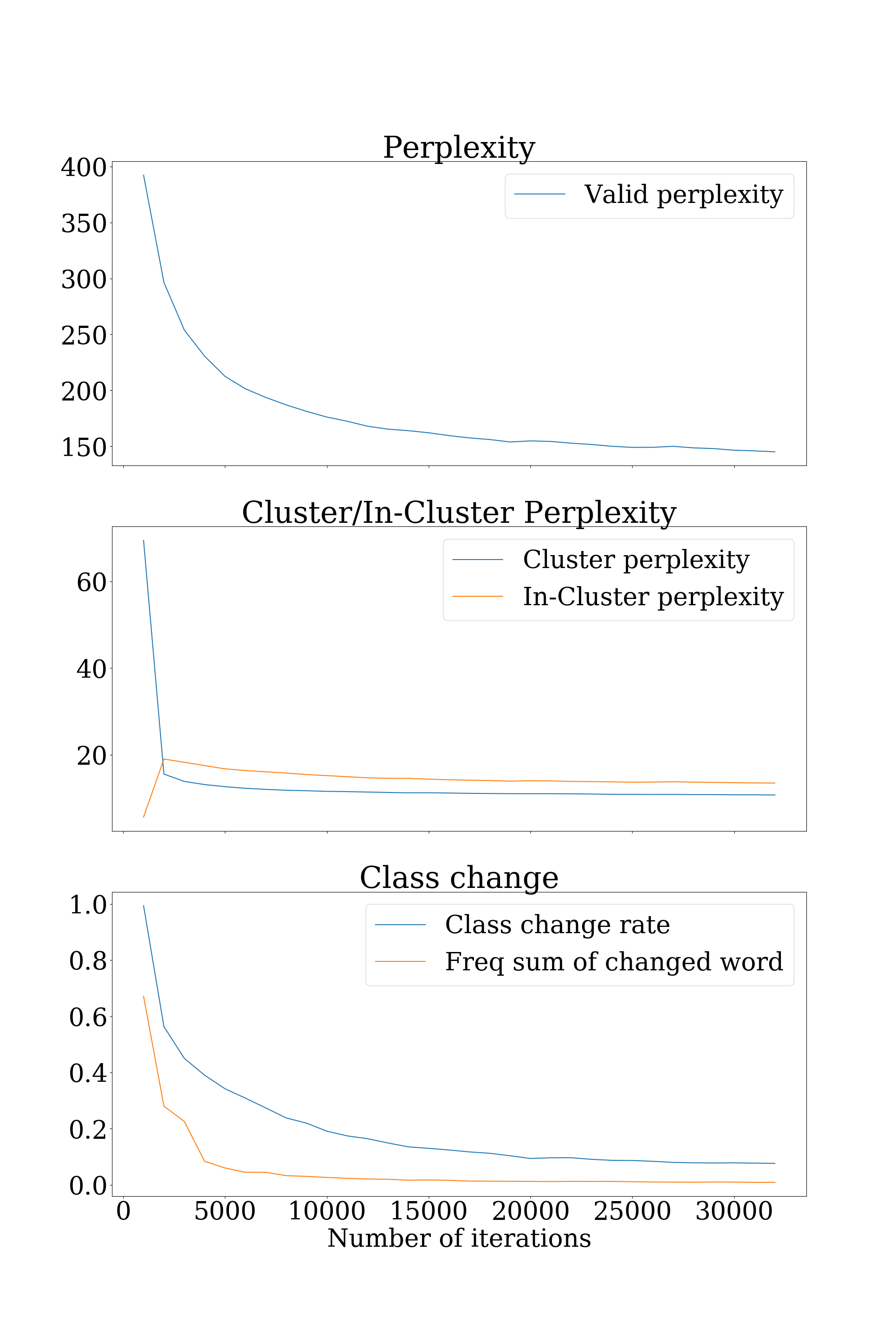}
	\caption{Perplexity, cluster perplexity, and in-cluster perplexity on the development set, while training on the text8 dataset. The "Class change rate" line shows the percentage of words that changed cluster after the cluster update algorithm. The "Freq sum of changed word" line shows term frequency sum of words that changed cluster.}
	\label{fig_frq}
\end{figure}

Table \ref{text8_result} shows results on the text8 dataset. Our approach provides the best perplexity among all approximation approaches, nearly performing as well as a full softmax. Table \ref{text8_time} shows that our approach is almost 4 times faster than a normal softmax, with the
speed-up continuing to increase as the vocabulary size increases.

Figure \ref{fig_frq} monitors learning process of our approach. At the beginning of training, we observe a high cluster perplexity, and very low in-cluster perplexity. Because we initialize clusters randomly, our model has difficulties to predict cluster given the preceding sequence context of the target word. 
As training continues, our cluster update algorithm 
considers the cluster word assignments based on the distribution given by Equation \ref{classprob_update}, resulting in the cluster perplexity decreasing rapidly.
In contrast, the in-cluster perplexity first increases and then decreases slowly. Because our approach assigns similar words into the same cluster that are difficult to distinguish,
the model has to explicitly learn small differences between words that share a similar context.
In the end, the model reaches a balance between cluster and in-cluster perplexity.

\begin{table*}[t]
	\center
  	\caption{Word clustering examples. Each line is a subset of words that belong to one cluster learn from the text8 corpus.}
    \label{cluster_example}
    \begin{tabular}{c|l}
    \hline
    Cluster & Words in same cluster \\
    \hline
    1 & be	have	use	do	include	make	become	support	take	show	play	change	run \\
   	\hline
    2 & transmitted     acclaimed       pressed shipped stolen  swept   marketed        contested       blamed judged \\
    \hline
    3 & delivering	selecting	issuing	stealing	imposing	lowering	asserting	supplying	regulating \\
    \hline
    4 & kn      mw      dy      volts   cubes   volt    kcal    mev     ohm     bhp     o    unces  megabytes \\
    \hline
    5 & empire  catholic        romans  byzantine       rulers  emperors        conquest        catholics       kingdoms  catholicism \\
    \hline
    6 & actor author writer singer actress director composer poet musician artist politician bishop \\
    \hline
    7 & zeus	achilles	venus	leto	heracles	saul	ptolemy	hera	aphrodite	beowulf	ajax	athena	caligula \\
    \hline
    8 & iraq	texas	afghanistan	sweden	boston	hungary	brazil	iran	wales	michigan	denmark	virginia \\
    \hline
    \end{tabular}
\end{table*}

Table \ref{cluster_example} shows some examples of words that belong to the same cluster. We observe a strong syntactical similarity between these words, and semantic closeness. These examples show that our cluster update algorithm is capable of placing words with similar context into the same cluster. It is interesting to see that the unsupervised approach could learn word cluster with clear meaning. 

\section{Abstractive Sentence Summarization Experiment}
Summarization is an important challenge in natural language understanding. The aim is to produce a condensed representation of an input text that captures the core meaning of the original. 

Given a vocabulary $\mathbf{V}$ and a sequence of $M$ words $x_1, ..., x_M \in \mathbf{V}$, a summarizer takes $\mathbf{x}$ as input and outputs a shortened sentence $\mathbf{y}$ of length $N < M$. Assuming that words in the output sentence also belong to $\mathbf{V}$, we can express the output as $y_1, ..., y_N \in \mathbf{V}$. The output sentence is then called a summary of the input sentence. Thus, the probability distribution of the summary can be defined as:
\begin{equation}
	P(y_1,...,y_N) = \prod_{t=1}^N P(y_t|\mathbf{x}, y_1,...,y_{t-1})
\end{equation}
For an extractive summarization, the probability distribution $P(y_t|\mathbf{x}, y_1,...,y_{t-1})$ is on the set of input words, while for an abstractive summarization the distribution is on the entire vocabulary. In this experiment, we focus on the abstractive summarization task, which is more difficult and computationally expensive.

\subsection{Dataset}
We trained our model on the Gigaword5 dataset \cite{napoles2012annotated}. This dataset was generated by pairing the headline of each article with its first sentence to create a source-compression pair. \cite{rush2015neural} provided scripts to filter out outliers, resulting in roughly 3.8M training pairs, a 400K validation set, and a 400K test set. We use the most frequent 69k words in the title as input and output vocabulary, which correspond to the decoder vocabulary size used in \cite{rush2015neural}. Out of vocabulary words are represented with a symbol $<unk>$.

We evaluate our method on both the standard DUC-2004 dataset and the single reference Gigaword5 test set. The DUC corpus\footnote{http://duc.nist.gov/duc2004/tasks.html} 2004 corpus consists of 500 documents, each having 4 human generated reference titles. Evaluation of this dataset uses the limited-length Rouge Recall at 75 bytes on DUC validation and test sets. In our work, we simply run the models trained on Gigaword corpus as they are,  without tuning them on the DUC validation set. The only change we have made to the decoder is to suppress the model from emitting the end-of-summary tag, and forced it to emit exactly 30 words for every summary. \cite{rush2015neural} provides a random sampled 2000 title-headline pair as test set. We acquired the exact test sample used by them to make a precise comparison of our models with theirs. Like \mbox{\cite{nallapati2016abstractive}} and \mbox{\cite{chopra2016abstractive}}, we use the full length F1 variant of Rouge\footnote{http://www.berouge.com/Pages/default.aspx} to evaluate our system.

\subsection{Implementation}
In this experiment, we use the standard Encoder-Decoder with attention architecture. Both encoder and decoder a consists of a single layer uni-direction LSTM model, and an attention mechanism over the source-hidden states and a softmax layer to output a distribution probability over an output vocabulary. 
The hidden state dimension and input embedding are both set to 512. All parameters are regularized with weight decay ($\lambda=10^{-6}$). Batch size is 128. We use Adam \cite{kingma2014adam} with a learning rate of 0.001. No dropout or gradient clipping is used. At decode time, we use beamsearch of size 5 to generate summary. The maximum length of output summary is limited to 30.

For our approach, we learn word clusters through training a language model on the titles of our training dataset. We then use these clusters as the fixed structure for hierarchical softmax in the summarization model.

\subsection{Baseline methods}
We compared the performance of our model with state-of-the-art models that are trained with teacher forcing and cross-entropy loss, including: (1) TOPIARY \cite{zajic2004bbn}, (2) ABS+ \cite{rush2015neural}, (3) RAS-Elman \cite{chopra2016abstractive}, and (4) words-1vk5k-1sent \cite{nallapati2016abstractive}. We also include our implementation of normal softmax as a baseline method. There are some newly proposed summarization models that come up with different type of loss function including the reconstructive loss function \cite{miao2016language}, and the minimum risk training (MRT) loss \cite{shen2016neural}. We did not compare with these methods, since this experiment is focused on evaluating our approach against other softmax-based approaches under similar implementations and learning settings.

\subsection{Experiment Results}
\begin{table}[t]
  \caption{Experiment results on DUC2004 testset}
  \label{DUC_result}
  \begin{tabular}{c|c|c|c }
     & $R1_{R}$ & $R2_{R}$ & $RL_{R}$ \\
    \hline
    TOPIARY & 25.16 & 6.46 & 20.12 \\
    ABS+ & 28.18 & 8.49 & 23.81 \\
    RAS-Elman & 28.97 & 8.26 & 24.06 \\
    words-1vk5k-1sent & 28.61 & 9.42 & 25.24 \\
    \hline
    Full softmax & 27.77 & 9.01 & 24.36 \\
    Self-organized HSM & \textbf{29.20} & \textbf{9.62} & \textbf{25.65} \\
  \end{tabular}
\end{table}

\begin{table}[t]
  \caption{Experiment results on Gigawords testset}
  \label{Gigawords_result}
  \begin{tabular}{c|c|c|c }
     & $R1_{F1}$ & $R2_{F1}$ & $RL_{F1}$ \\
    \hline
    ABS+ & 29.76 & 11.88 & 26.42 \\
    RAS-Elman & 33.78 & 15.97 & 31.15 \\
    words-1vk5k-1sent & 33.17 & \textbf{16.02} & 30.98 \\
    \hline
    Full softmax & 33.54 & 15.49 & 31.52 \\
    Self-organized HSM & \textbf{34.07} & 16.00 & \textbf{31.95} \\
  \end{tabular}
\end{table}

\begin{figure}[!htp]
  \scriptsize
  \begin{framed}    
\textbf{I(1):} china is expected to become the world 's largest market for motorcycles by \#\#\#\# , with annual demand topping \#\# million units , according to a report published wednesday . \\
\textbf{G:} china to become world 's largest motorcycle market \\
\textbf{S:} china to become world 's largest motorcycles \\
\textbf{H:} china to become world 's largest motorcycle market \\

\textbf{I(2):} eleven opposition parties went to tanzania 's high court wednesday to seek a ruling nullifying the east african nation 's tanzania 's first multi-party elections . \\
\textbf{G:} opposition seeks nullification of elections \\
\textbf{S:} \#\# opposition parties go to tanzania court \\
\textbf{H:} tanzanian opposition seeks ruling on tanzanian elections \\

\textbf{I(3):} india has secured contracts from egypt and syria to supply spare parts for mig fighters , breaking russia 's virtual monopoly , officials here said wednesday . \\
\textbf{G:} india beats russia to supply mig spares to egypt syria \\
\textbf{S:} india secures contracts from egypt syria  \\
\textbf{H:} india to supply parts for mig fighters  \\

\textbf{I(4):} sri lankan security forces wednesday geared up for the `` bloodiest fighting '' yet in their bid to capture the tamil tiger headquarters at jaffna town , amid intense resistance from the rebels , military officials said . \\
\textbf{G:} sri lanka braces for bloodiest battle tigers deny fleeing by amal jayasinghe \\
\textbf{S:} sri lanka gears up for jaffna fighting \\
\textbf{H:} sri lankan security forces gear up for fighting \\

\textbf{I(5):} japanese prime minister tomiichi murayama told us defense secretary william perry here wednesday he would uphold bilateral security ties despite strong opposition to the us bases in okinawa . \\
\textbf{G:} murayama vows to uphold security ties with us \\
\textbf{S:} murayama vows to defend bilateral security ties \\
\textbf{H:} murayama vows to uphold security ties with us \\

\textbf{I(6):} russian defence minister pavel grachev left for moscow on wednesday after winding up a three-day official visit to greece during which he signed a military cooperation accord with his counterpart gerassimos arsenis . \\
\textbf{G:} grachev ends three-day visit to greece \\
\textbf{S:} russian minister leaves for moscow \\
\textbf{H:} grachev leaves greece for moscow \\

\textbf{I(7):} an explosion in iraq's restive northeastern province of diyala killed two us soldiers and wounded two more , the military reported monday . \\
\textbf{G:} two us soldiers killed in iraq blast december toll \#\#\# \\
\textbf{S:}  two us soldiers killed in iraq \\
\textbf{H:} two us soldiers killed two wounded in iraq \\

  \end{framed}
  \vspace*{-0.5cm}
  \caption{\small \label{examples} Example sentence summaries produced on Gigaword. \textbf{I} is the input, \textbf{G} is the true headline, \textbf{S} is the full softmax, and \textbf{H} is the self-organized HSM.}
\end{figure}

\begin{table}[t]
	\center
	\caption{Training time on Gigaword5 training set}
	\begin{tabular}{c|c}
	& Training time \\
    \hline
    Full softmax & 210 min \\
    Self-organized HSM & 63 min
	\end{tabular}
    \label{giga_time}
\end{table}

Table \ref{DUC_result} and Table \ref{Gigawords_result} show results of our approach comparing the different methods. Our approach not only outperforms the full softmax method, but also outperforms state-of-the-art methods on most of evaluation metrics. Table \ref{giga_time} also shows that our approach is 3 times faster than standard full softmax, which is widely used in all kinds of different summarization models. Figure \ref{examples} present examples of summaries generated by self-organized HSM in comparing with true headline and full softmax outputs.

As the word cluster learned on Gigaword corpus shows similar in-cluster syntactical similarity and semantic closeness, we suggest that this hierarchical structure decomposes the difficult word generation task into 2 easier tasks. The first task is to decide correct syntactical role for next word, and the second task is to find semantically correct words in a subset of the vocabulary.

\section{Conclusion}
In this paper we have proposed a new self organizing variant of the hierarchical softmax. We observe that this approach can achieve the same performance as a full softmax approach for language modelling, and even better performance for sentence summarization. Our approximation approach is also as efficient as other hierarchical softmax approximation techniques. In particular in our experiments we observe that our self-organized HSM is at least 3 times faster than a full softmax approach.

Our approach yields self organized word clusters which are influenced by the context of words.
Examining word clusters produced by or approach reveals that our method groups words according to their syntactical role and semantic similarity. These results are appealing in that we have obtained a certain level of understanding of grammar rules without explicit part-of-speech tagged input. We therefore think that the use of this approach shows promise for other NLP tasks, including machine translation and natural language generation.

\bibliography{main}
\bibliographystyle{emnlp_natbib}

\end{document}